\documentclass{article}
\usepackage[utf8]{inputenc}
\usepackage{amsfonts,amsmath}
\usepackage{mdframed}
\usepackage{graphicx}
\usepackage{hyperref}

\usepackage{tikz}
\usetikzlibrary{arrows.meta,,positioning}

\usepackage{caption}
\usepackage[ruled,boxed,linesnumbered,commentsnumbered,longend]{algorithm2e}

\usepackage{amsthm}
\theoremstyle{definition}

\usepackage{multirow}
\usepackage{breqn}

\def\old#1{}

\title{Reinforcement Learning Methods\\ for Wordle: A POMDP/Adaptive Control Approach}
\author{Siddhant Bhambri,  Amrita Bhattacharjee, \\and Dimitri Bertsekas\thanks{School for Computing and Augmented Intelligence, Arizona State University}}
\date{\{sbhambr1, abhatt43, dbertsek\}@asu.edu}

\begin{document}
\maketitle

\begin{abstract}
\label{abstract}
In this paper we address the solution of the popular Wordle puzzle, using new reinforcement learning methods, which apply more generally to  adaptive control of dynamic systems and to classes of Partially Observable Markov Decision Process (POMDP) problems. These methods are based on approximation in value space and the rollout approach, admit a straightforward implementation, and provide improved performance over various heuristic approaches. For the Wordle puzzle, they yield on-line solution strategies that are very close to optimal at relatively modest computational cost. Our methods are viable for more complex versions of Wordle and related search problems, for which an optimal strategy would be impossible to compute. They are also applicable to a wide range of adaptive sequential decision problems that involve an unknown or frequently changing  environment whose parameters are estimated on-line.
\end{abstract}


\newpage

\section{Introduction}
\label{sec1:introduction}

In this paper, we discuss a Reinforcement Learning (RL) approach towards a class of sequential decision problems, exemplified for the popular Wordle puzzle that appears daily in the New York Times. Wordle involves a list of 5-letter mystery words, which is a subset of a larger list of guess words. A word is selected at random from the  mystery list, and the objective is to find that word by sequentially selecting no more than six words from the guess list. Each guess word selection provides information about the letters contained in the hidden mystery word according to a given set of rules, which involves color coding of letters shared by the guess word and the mystery word. 

We will adopt a more general point of view, by considering a broad class of problems that include Wordle as a special case. In particular, the problems that we consider include sequential search situations, where the objective is to guess correctly an unknown object from a given finite set of objects (the set of mystery words in the Wordle context), by using a sequence of decisions from a finite set (the set of guess words in Wordle), which result in a sequence of corresponding observations (the information outcomes of the guesses in Wordle). We aim to minimize some cost function, such as the expected number of observations required to determine the unknown object. 

Within the search context just described, some basic information theory concepts are relevant, which have already been applied to Wordle, and are important for our methodology. In particular, consider a random variable $\Theta$ that can take a finite number of values $\theta^1,\ldots,\theta^m$ with given probabilities. Suppose that we want to estimate $\Theta$ and to this end, we select one out of a finite set of observations $Z_u$, where $u$ is a parameter that takes values in  a finite set $U$. Each $Z_u$ is a random variable that can take a finite number of values $z^1,\ldots,z^n$, and provides information about the true value of $\Theta$ through the conditional probability distribution $P_{\Theta\mid Z_u}$, which is assumed known for each $u\in U$. An information theoretic approach suggests choosing $u$ to be  one that results in maximum entropy reduction (or maximizes the information gain), as measured by notions of entropy. In particular, the (a priori) entropy of $\Theta$ is given by
\begin{equation}
\label{eq:entropy}
    H(\Theta)=-\sum_{i=1}^m p(\theta^i)\log\big(p(\theta^i)\big),
\end{equation}
where $p(\theta^i)$ is the a priori probability that $\Theta$ has the value $\theta^i$. The a posteriori entropy of $\Theta$ given $Z_u$ is given by
\begin{equation}
\label{eq:posterior_entropy}
    H(\Theta\mid Z_u)=-\sum_{j=1}^np(z^j)\sum_{i=1}^m p(\theta^i\mid Z_u=z^j)\log\big(p(\theta^i\mid Z_u=z^j)\big),    
\end{equation}
where $p(z^j)$ is the probability that $Z_u$ takes the value $z^j$ and $p(\theta^i\mid Z_u=z^j)$ is the conditional probability that $\Theta=\theta^i$ given that $Z_u$ has taken the value $z^j$.
The entropy reduction (or information gain) provided by a choice $u\in U$ is the function of $u$ given by
\begin{equation}
\label{eq:entropy_reduction}
    H(\Theta)-H(\Theta\mid Z_u),    
\end{equation}
and the  information theoretic approach suggests selecting $u\in U$ that maximizes the above expression, or equivalently minimizes the conditional entropy $H(\Theta\mid Z_u)$.

Thus, if a single guess word were allowed within the Wordle context, it would be reasonable to select the one that results in maximum information gain. In the real case where multiple (namely six) guess words are allowed, it makes sense to apply the maximum information gain approach sequentially, i.e., after the results of a guess become known, obtain the resulting conditional distribution of $\Theta$, evaluate the information gain that corresponds to each candidate guess word, and select as next guess word one that maximizes the information gain. 

The information gain approach for sequential guess word selection has been proposed by Sanderson through his popular 3Blue1Brown\footnote{\url{https://www.youtube.com/c/3blue1brown}} channel, and  has near-optimal performance for the ``standard" form of the puzzle [within 5.5\% in the ``easy" mode and 2.8\% in the ``hard" mode, assuming the optimal opening word selection (\texttt{salet)}]. It may also be viewed as a sub-optimal/heuristic policy for an underlying dynamic programming (DP) problem. 

Our solution methodology for Wordle is based on DP and relies for the most part on the maximum information gain policy. It improves this policy by using a rollout approach, which amounts to a single policy iteration (i.e., start from the maximum information gain policy to obtain a rollout policy through a policy improvement operation). Indeed we will show computationally that the rollout policy performs substantially better than the maximum information gain policy, and is very close to optimal (it performs within 0.4\% of the optimal assuming the best opening word selection \texttt{salet}). Our rollout solution methodology can also be easily adapted for use with other heuristics, and similarly improves substantially their performance, based on computational experimentation with two heuristics (``most rapid decrease" and ``greatest expected probability") that have been suggested in the literature [Sho22]; see Sections \ref{sec2:formulation} and \ref{sec3:expts&results}. 

The performance improvement just described can be explained through the interpretation of our method as a step of Newton's method for solving an underlying Bellman equation, which starting from a cost function approximation provided by the heuristic policy, yields a rollout policy performance that is close to optimal. This interpretation is the principal theme of the recent monograph [Ber22] by one of the authors, and its more mathematical antecedent textbook [Ber20]. It is valid in great generality for approximation in value space schemes, well beyond the context of the present paper.

In this paper we will explain and develop our DP-based rollout approach within a generalized context that has applications  beyond sequential search. In particular, we will consider a broad class of adaptive control problems, which  involve a dynamic system with a model containing unknown parameters, which are directly or indirectly estimated using sequential observations, as the system is being controlled. We cast these adaptive control problems as POMDP; this is an approach that dates from the 1970s, and will be explained in Section \ref{sec4:pomdp}. We then use our adaptive control/POMDP approach as a starting point for the application of RL methods, based on approximation in value space and rollout (an extensive account of such methods is given in RL books, such as [BeT96], [SuB18], [Ber19], and particularly the books [Ber20] and [Ber22], which contain many earlier references). Aside from its effectiveness within the Wordle context, our approach highlights the benefits of the synergy between the artificial intelligence/RL, and the decision and control methodologies. In particular, in RL one is often faced with partially known environments, which are progressively learned through sequential observations, for the purpose of enhancing the selections of future actions. The adaptive control/POMDP framework of this paper also aims to address similar situations.

The paper is organized as follows. In Section \ref{sec2:formulation}, we introduce this formulation informally and describe our rollout approach within the context of Wordle. In Section \ref{sec3:expts&results}, we discuss our implementations and the corresponding computational results. In  Section \ref{sec4:pomdp}, we explain in more general mathematical terms our approach, within a deterministic adaptive control context. In Section \ref{sec5:extensions}, we discuss various extensions of our methodology, both within the context of Wordle (e.g., more challenging versions of the puzzle), and within the context of sequential search and adaptive control (including extensions that involve stochastic systems).

\section{Playing Wordle: Problem Formulation and Rollout Algorithms}
\label{sec2:formulation}

We provide a brief introduction to the Wordle puzzle, which was created by Josh Wardle [War22], and published online in January, 2022. It is reputedly played by millions of people daily.\footnote{\url{https://fictionhorizon.com/how-many-people-play-wordle}} In Wordle, players have to guess a five-letter word in six attempts. The hidden word, which we will refer to as the ``mystery word" in this paper, is chosen every day from a list of 2,315 words according to some distribution and posted on The New York Times website. In published studies, including the reports of optimal and suboptimal computational results, as well as the present paper, the distribution is assumed to be uniform. This list, referred to as the ``mystery list," is known to the public. Each guess attempt consists of a word chosen from a ``guess list" of 12,972 words, also known to the public, and provides information about the letter in each of the five positions of the mystery word.

\begin{figure}[ht]
\begin{center}
\centerline{\fbox{\includegraphics[width=0.75\columnwidth]{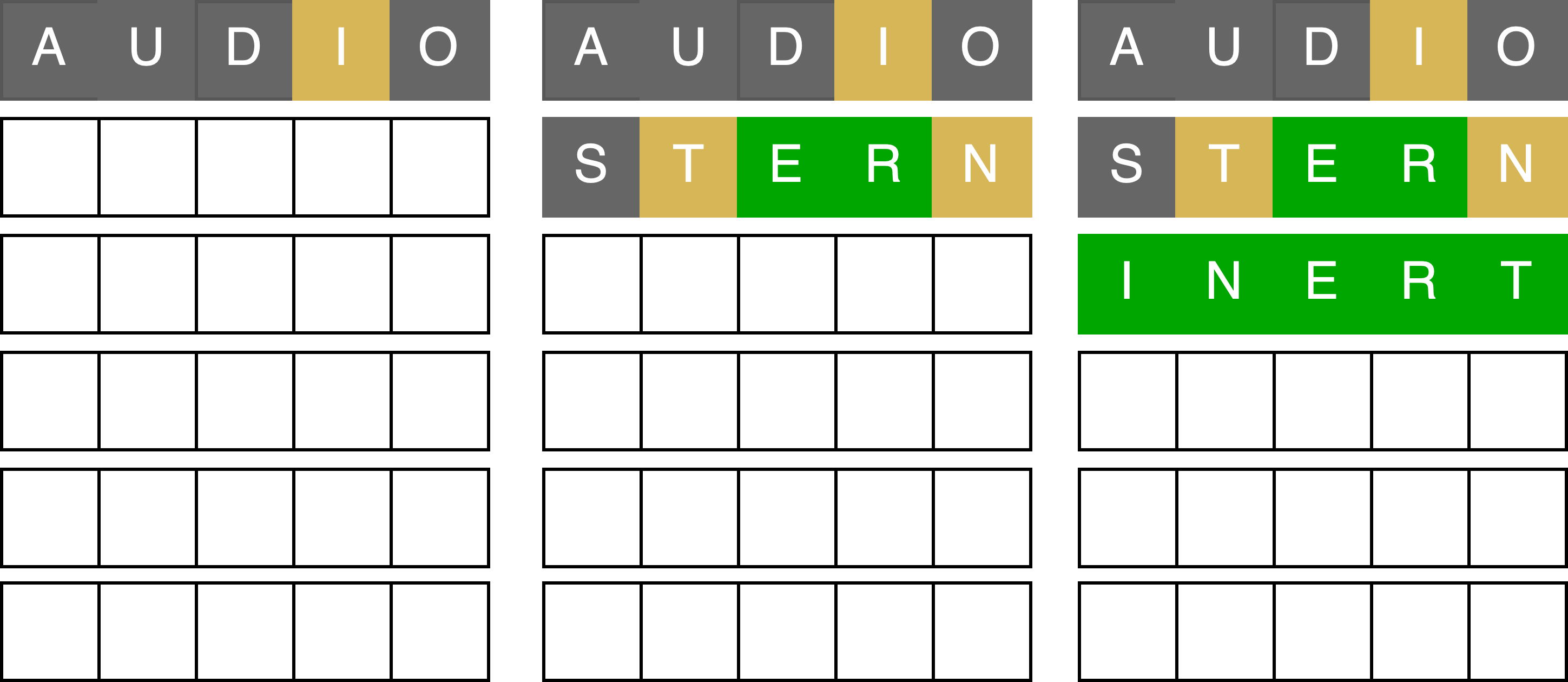}}}
\caption{Playing the Wordle game in easy mode, where the user has the flexibility to choose any word as the next guess word.}
\label{fig:wordle_easy}
\end{center}
\end{figure}

\begin{figure}[ht]
\begin{center}
\centerline{\fbox{\includegraphics[width=0.75\columnwidth]{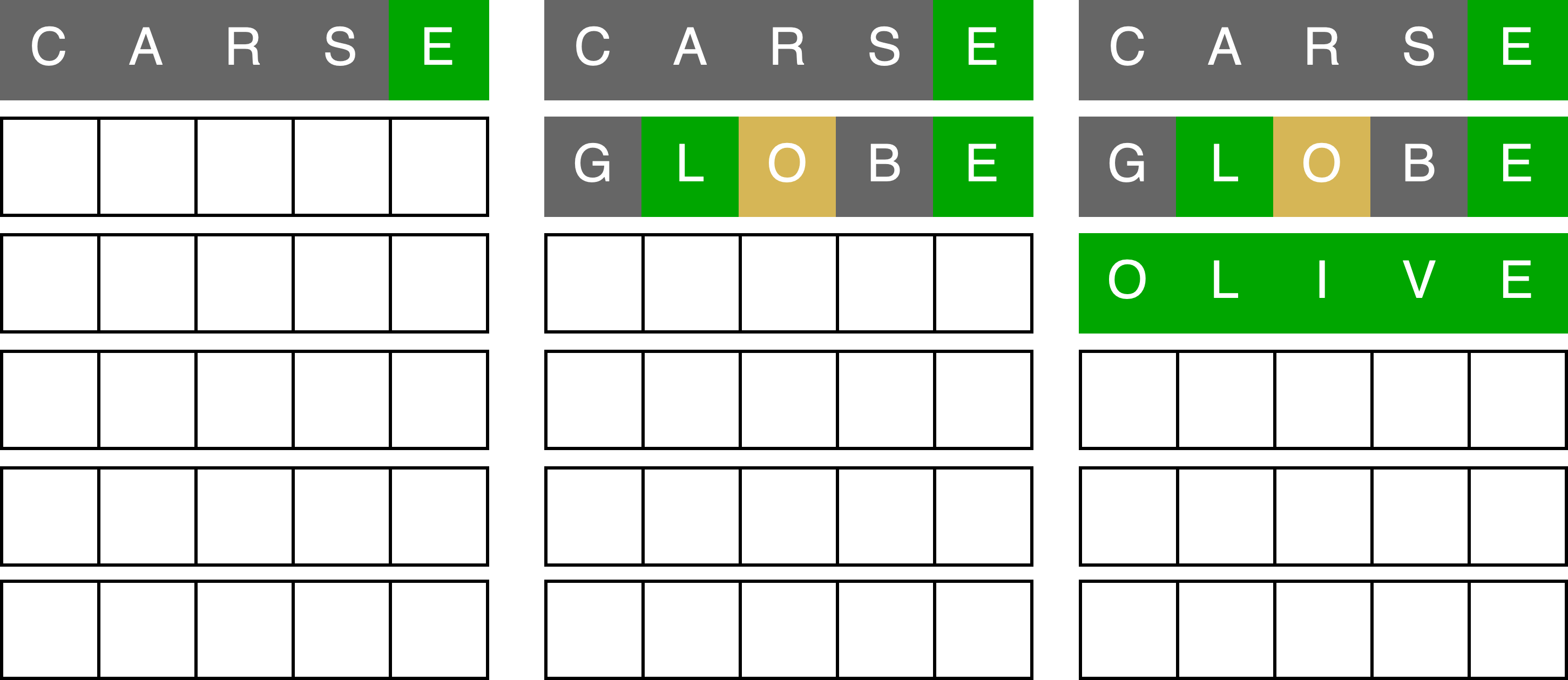}}}
\caption{Playing the Wordle game in hard mode, where the user is constrained to use the letters marked as ``yellow" and the letters marked as ``green" that have to be at the same position as in the previous guess word. For example, here, as we get ``E" as green when we play \texttt{CARSE}, we need to use only the words that end with an ``E", and so on.}
\label{fig:wordle_hard}
\end{center}
\end{figure}

There are two modes for playing the game, the ``easy" mode whereby the choice of each guess word is unrestricted within the guess list, and the ``hard mode" whereby the choice of each guess word depends on the outcomes of the preceding choices according to certain rules. For examples of solving the puzzle  in easy mode and hard mode, see Figs.\ \ref{fig:wordle_easy} and \ref{fig:wordle_hard}, respectively. We refer the readers to the internet literature\footnote{\url{https://www.nytimes.com/games/wordle/index.html}} for a detailed description of the rules of the puzzle.

Wordle has attracted the attention of quite a few scientists, and there is a growing body of analysis and algorithmic development from mathematical and computer science perspectives. Most of the discussions have focused on attempts to find the optimal strategy for guess word selection, or to propose good sub-optimal strategies. In particular, a widely known work by Selby [Sel22] was the first to implement optimal strategies for playing Wordle, and gave the corresponding optimal scores. These scores are 3.4212 attempts on the average to solve the puzzle in easy mode, and 3.5084 attempts on the average to solve the puzzle in hard mode, with the optimal opening word being \texttt{salet} in both modes of the game. The averages correspond to a mystery word chosen randomly from within the mystery list according to a uniform probability distribution. A subsequent paper by Bertsimas and Paskov [BeP22] has verified these optimal scores, using a DP-based solution method, which is apparently similar to the one used by Selby (the paper does not include a methodological comparison or reference to [Sel22]). The paper also reports that the puzzle took ``days to solve via an efficient C++ implementation of the algorithm, parallelized across a 64-core computer." Note that once this long DP computation is done, the on-line optimal solution of the puzzle for a single mystery word is extremely fast, using essentially table lookup (unless of course some changes are made to the mystery list, the guess list, or the rules of the puzzle, in which case the DP computation has to be repeated). On the other hand, the optimal DP solution of simple variants of the puzzle, such as for example using a known non-uniform distribution to choose the mystery word from within the mystery list, may be completely outside the realm of practical feasibility.

Several works have used some kind of heuristic strategy to approximate the optimal on-line guess word selection. The work of Sanderson, noted in Section \ref{sec1:introduction},  popularized the use of the information theory perspective and the maximum information gain heuristic to solve the puzzle sub-optimally. We have mostly focused on this heuristic for use in our rollout approach, which is described mathematically in Section \ref{sec4:pomdp}, and somewhat informally later in the present section. 

Two alternative heuristics for solving the puzzle on-line, which we have tested and incorporated into our rollout approach, are the ``most rapid decrease" algorithm and the ``greatest expected probability" algorithm,  proposed in the paper by Short [Sho22]. The most rapid decrease algorithm is related to the information theoretic perspective that we mentioned in Section \ref{sec1:introduction}. It aims to reduce the entropy of the mystery list by minimizing its size with each successive guess word choice. The greatest expected probability algorithm is based on choosing a guess ``word that leads to the greatest expected probability that you could randomly choose the correct answer on the next round of play," according to the description of [Sho22]. We have used an implementation of this heuristic, as best as we could understand it. Additional heuristic strategies have been proposed by Bonthron [Bon22] and Silva [Sil22]. We also mention that RL solution methods such as Deep-Q Learning and Advantage Actor Critic were tested by [HoA22], with considerably worse results to the maximum information gain heuristic. Apart from sub-optimal strategies aimed at solving the puzzle on-line, there has also been  work aimed at obtaining theoretical complexity guarantees; see Lokshtanov and Subercaseaux [LoS22], and Rosenbaum [Ros22] for representative works. 

\subsection{Rollout Algorithm for Wordle}

\begin{figure}[ht]
\begin{center}
\centerline{\fbox{\includegraphics[width=0.65\columnwidth]{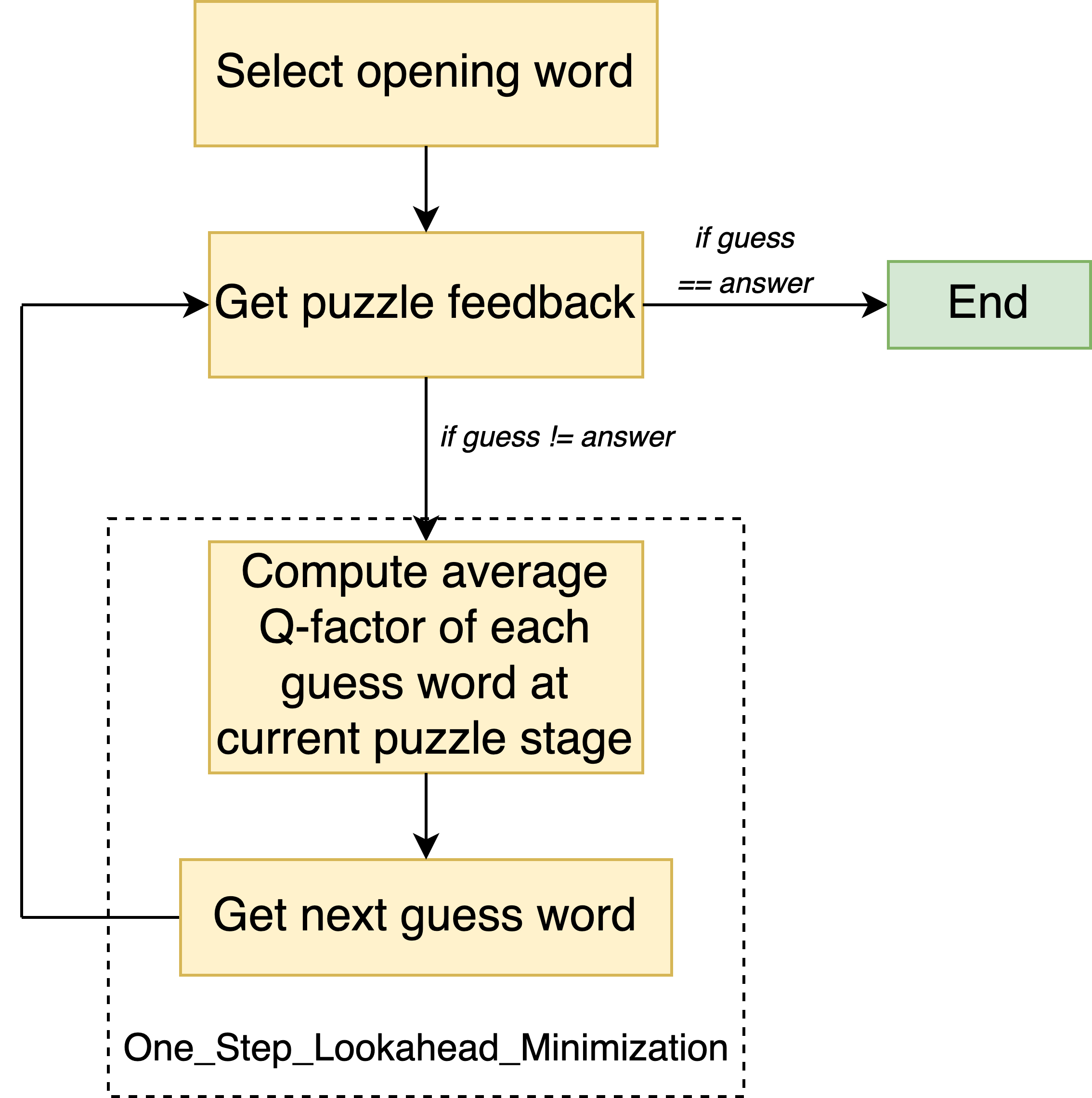}}}
\caption{Pseudocode structure for executing Wordle; see Algorithm 1 description.}
\label{fig:wordle_block}
\end{center}
\end{figure}

We will now describe somewhat informally our rollout approach for Wordle. Assume that we have a heuristic algorithm, denoted by  $H$, which selects a guess word in response to the information obtained in response to a preceding set of guess word selections. We refer to $H$ as the {\it base heuristic\/}. The rollout algorithm that uses $H$ as a base heuristic aims to improve the performance of $H$. 

At a typical stage of the rollout algorithm, say stage $k$, we know the {\it current mystery list} $M_k$, which is defined as the subset of mystery words that are possible based on the information received in response to the preceding $k-1$ guess word selections (this is the narrowed-down subset of the original mystery list that is consistent with the feedback earlier obtained). We also know the {\it current guess list} $U_k$, defined  as the subset of words in the original guess list $U$ that are allowable for our next selection based on the information received in response to the preceding $k-1$ guess word selections. Note, that in the easy mode we have $U_k=U$, but in the hard mode $U_k$ may be a strict subset of $U$. The rollout algorithm chooses the $k$th guess word $\tilde u_k\in U_k$, the puzzle responds with some information regarding the identity and position of some letters within the mystery word according to the rules of the puzzle, the new mystery and guess lists, $M_{k+1}$ and $U_{k+1}$, are formed, and the process is repeated until the mystery word is identified. The method for choosing $\tilde u_k$, the guess word for the current stage $k$, will now be described.

For each pair $(m,u)$, consisting of a possible mystery word $m\in M_k$ and guess word $u\in U_k$, we calculate $Q_k(m,u)$, the {\it Q-factor of $u$, conditioned on $m$ being the true mystery word\/}. This is the number of guesses required to find $m$, assuming that we select $u$ as our first guess, and then we select the subsequent guess words using the base heuristic $H$. This Q-factor can be simply computed by simulating the base heuristic forward from stage $k+1$, knowing $m$ and the $k$th stage selection $u$. We also compute for each $u\in U_k$, the  {\it average Q-factor of $u$\/}, denoted  $\hat Q_k(u)$, as
\begin{equation}
\label{eq:qfactor}
\hat Q_k(u)={\frac{1}{\hbox{$|M_k|$}}}\sum_{m\in M_k}Q_k(m,u),
\end{equation}
where, $|M_k|$ denotes the cardinality of the mystery word list $M_k$.
The rollout algorithm then selects a guess word $\tilde u_k$ whose average Q-factor is minimal:
\begin{equation}
\label{eq:qfactormin}
\tilde u_k\in\arg\min_{u\in U_k}\hat Q_k(u).
\end{equation}
This algorithmic description is consistent with the more formal mathematical description that will be given in Section 4.

In our implementation, we are not calculating the Q-factors $Q_k(m,u)$ for all $u\in U_k$, since $U_k$ can be very large (as large as 12,972 words). Instead, given $M_k$ and $u\in U_k$, we compute a table of scores provided by the base heuristic for just the current stage (i.e., in the case of maximum information gain, the information gain provided by $u$, when the mystery list is $M_k$). Based on the results, we select a subset $\hat U_k$ of top performing words $u\in U_k$. In other words, we  replace the set $U_k$ in the Q-factor minimization (\ref{eq:qfactormin}) with a subset $\hat U_k\subset U_k$ consisting of a number of best performing words according to a single-stage version of the base heuristic. In our experiments, $\hat U_k$ consists of the top 10 guess words from $U_k$; increasing the number of top guess words to 100 yielded minor performance improvements at the expense of substantially longer computation. 

A block diagram for the rollout algorithm is given in Fig.\ \ref{fig:wordle_block}. An instance of the puzzle is started by choosing a randomly chosen mystery word as the answer. We then select one of the standard opening words, known to yield good results on the average, and based on the feedback received from the puzzle, we proceed. At each stage, if the answer has not been found, the top $10$ guess words are selected, as described above, the corresponding average Q-factors are computed according to Eq.\ (\ref{eq:qfactor}), and the guess word with minimal average Q-factor is selected.  A pseudocode describing the two components of this algorithm is given in Algorithm \ref{algo-rollout}.

\begin{algorithm*}
\SetAlgoLined
\LinesNumbered
\SetKwFunction{FMain}{One_Step_Lookahead_Minimization}
\SetKwProg{Fn}{Function}{:}{}

\Fn{\FMain{$U_i, M_i, priors, ents$}}{
$ents \leftarrow get\_entropy\_scores(U_i, M_i, priors)$\;
$max\_indices \leftarrow argmax(ents)$\;
$top\_choices \leftarrow U_i[max\_indices][:top_k]$\;
$top\_choices\_Q \leftarrow [\quad]$\;
\For{$choice \in top\_choices$}{
$choice\_q\_factors \leftarrow [\quad]$\;
\For{$m \in M_i$}{

$guess \leftarrow choice$\;
\While{$guess \neq m$}{
$guess \leftarrow$ guess word with maximum information gain for current mystery list\;
$score \leftarrow score + 1$\;}
$choice\_q\_factors.append(score)$\;
}
$mean\_q \leftarrow \frac{1}{|M_i|}\sum(choice\_q\_factors)$\;
$top\_choices\_Q.append(mean\_q)$\;

}
$min\_index \leftarrow argmin(top\_choices\_Q)$\;
\KwRet $U_i[min\_index]$\;
}
\caption{$One\_Step\_Lookahead\_Minimization$ with Rollout (using MIG as the base heuristic).}
\label{algo-rollout}
\end{algorithm*}



\section{Experiments with Rollout and Computational Results}
\label{sec3:expts&results}

In this section, we describe our experiments and computational results using our rollout approach and the three base heuristics noted earlier. We focus primarily on the maximum information gain heuristic (MIG for short), but we have also tested the other two heuristics: most rapid decrease (MRD for short) and greatest expected probability (GEP for short) as discussed in [Sho22]. We have re-implemented the MRD and GEP heuristics, based on our best understanding from the original work in [Sho22], so our implementations may differ from the ones of [Sho22].\footnote{Actually, our implementation of MRD performs very well relative to the optimal. This is not true for GEP, which performs rather poorly. Remarkably, however, our rollout algorithm that uses GEP as its base heuristic, yields near-optimal performance; see Table 2.} Still, our implementations represent legitimate heuristics, so they are suitable for comparison with the corresponding rollout algorithm, which is our principal aim.

In summary, our tests show that our rollout approach improves substantially upon the performance of the  three heuristics. In particular, the rollout performance is very close to the optimal, as calculated in the papers [Sel22] and [BeP22], even in the case of the GEP base heuristic, whose performance is far from optimal. This is consistent with a general interpretation of rollout and approximation in value space methods as a single step of Newton's method for solving the Bellman equation associated with the underlying DP problem; cf.\ the books [Ber20] and [Ber22]. The role of the base heuristic is to provide the starting point for the Newton step, and apparently all three base heuristics provide starting points that are within the region of fast convergence of Newton's method.

\begin{table}[ht]
\caption{Results using maximum information gain (MIG) as base heuristic and with rollout. Missing entries for optimal score were not given in [Sel22] or [BeP22]. For easy mode, rollout comes within 0.4\% of the optimal score, compared with 5.5\% for MIG. For hard mode, rollout comes within 0.4\% of the optimal score, compared with 2.8\% for MIG.}
\label{tab:max_info_gain}
\resizebox{\textwidth}{!}{%
\begin{tabular}{|c|ccc|ccc|}
\hline
\multirow{2}{*}{\textbf{Opening Word}} & \multicolumn{3}{c|}{\textbf{Easy Mode}}                                                                                                                                                                                                 & \multicolumn{3}{c|}{\textbf{Hard Mode}}                                                                                                                                                                                                 \\ \cline{2-7} 
                                       & \multicolumn{1}{c|}{\textbf{\begin{tabular}[c]{@{}c@{}}MIG as \\ Base Heuristic\end{tabular}}} & \multicolumn{1}{c|}{\textbf{\begin{tabular}[c]{@{}c@{}}Rollout with\\ MIG as \\ Base Heuristic\end{tabular}}} & \textbf{Optimal Score} & \multicolumn{1}{c|}{\textbf{\begin{tabular}[c]{@{}c@{}}MIG as \\ Base Heuristic\end{tabular}}} & \multicolumn{1}{c|}{\textbf{\begin{tabular}[c]{@{}c@{}}Rollout with\\ MIG as \\ Base Heuristic\end{tabular}}} & \textbf{Optimal Score} \\ \hline
\textbf{salet}                         & \multicolumn{1}{c|}{3.6108}                                                                    & \multicolumn{1}{c|}{3.4345}                                                                                   & 3.4212                 & \multicolumn{1}{c|}{3.6078}                                                                    & \multicolumn{1}{c|}{3.5231}                                                                                   & 3.5084                 \\ \hline
\textbf{reast}                         & \multicolumn{1}{c|}{3.6}                                                                       & \multicolumn{1}{c|}{3.4462}                                                                                   & 3.4225                 & \multicolumn{1}{c|}{3.6181}                                                                    & \multicolumn{1}{c|}{3.53}                                                                                     & 3.5136                 \\ \hline
\textbf{crate}                         & \multicolumn{1}{c|}{3.6177}                                                                    & \multicolumn{1}{c|}{3.4414}                                                                                   & 3.4238                 & \multicolumn{1}{c|}{3.6289}                                                                    & \multicolumn{1}{c|}{3.5361}                                                                                   & 3.5175                 \\ \hline
\textbf{trace}                         & \multicolumn{1}{c|}{3.6069}                                                                    & \multicolumn{1}{c|}{3.4393}                                                                                   & 3.4238                 & \multicolumn{1}{c|}{3.6212}                                                                    & \multicolumn{1}{c|}{3.5266}                                                                                   & -                      \\ \hline
\textbf{slate}                         & \multicolumn{1}{c|}{3.6142}                                                                    & \multicolumn{1}{c|}{3.4362}                                                                                   & 3.4246                 & \multicolumn{1}{c|}{3.6129}                                                                    & \multicolumn{1}{c|}{3.5227}                                                                                   & -                      \\ \hline
\textbf{trape}                         & \multicolumn{1}{c|}{3.6319}                                                                    & \multicolumn{1}{c|}{3.4604}                                                                                   & 3.4454                 & \multicolumn{1}{c|}{3.6199}                                                                    & \multicolumn{1}{c|}{3.5356}                                                                                   & 3.5179                 \\ \hline
\textbf{slane}                         & \multicolumn{1}{c|}{3.6255}                                                                    & \multicolumn{1}{c|}{3.4444}                                                                                   & 3.4311                 & \multicolumn{1}{c|}{3.622}                                                                     & \multicolumn{1}{c|}{3.5378}                                                                                   & 3.5201                 \\ \hline
\textbf{prate}                         & \multicolumn{1}{c|}{3.6333}                                                                    & \multicolumn{1}{c|}{3.4535}                                                                                   & 3.4376                 & \multicolumn{1}{c|}{3.6173}                                                                    & \multicolumn{1}{c|}{3.5348}                                                                                   & 3.5210                 \\ \hline
\textbf{crane}                         & \multicolumn{1}{c|}{3.6091}                                                                    & \multicolumn{1}{c|}{3.4380}                                                                                   & 3.4255                 & \multicolumn{1}{c|}{3.6333}                                                                    & \multicolumn{1}{c|}{3.5374}                                                                                   & 3.5227                 \\ \hline
\textbf{carle}                         & \multicolumn{1}{c|}{3.6108}                                                                    & \multicolumn{1}{c|}{3.4419}                                                                                   & 3.4285                 & \multicolumn{1}{c|}{3.6384}                                                                    & \multicolumn{1}{c|}{3.5369}                                                                                   & 3.5261                 \\ \hline
\textbf{train}                         & \multicolumn{1}{c|}{3.6181}                                                                    & \multicolumn{1}{c|}{3.4622}                                                                                   & 3.4436                 & \multicolumn{1}{c|}{3.6216}                                                                    & \multicolumn{1}{c|}{3.5369}                                                                                   & 3.5248                 \\ \hline
\textbf{raise}                         & \multicolumn{1}{c|}{3.6389}                                                                    & \multicolumn{1}{c|}{3.4777}                                                                                   & 3.4618                 & \multicolumn{1}{c|}{3.6721}                                                                    & \multicolumn{1}{c|}{3.5866}                                                                                   & -                      \\ \hline
\textbf{clout}                         & \multicolumn{1}{c|}{3.6955}                                                                    & \multicolumn{1}{c|}{3.5248}                                                                                   & -                      & \multicolumn{1}{c|}{3.7123}                                                                    & \multicolumn{1}{c|}{3.6125}                                                                                   & -                      \\ \hline
\end{tabular}%
}
\end{table}

More specifically, we have evaluated the three heuristics and their use as base policies within the rollout approach for a selected set of opening words, which have been identified in earlier works as best or nearly best choices for initial guess selection. In particular, we have included the 2 opening words suggested as best in [Sil22], the top 5 words according to the optimal results obtained by [BeP22], and the top 10 words from the optimal results given in [Sel22]. In our experiments, we use the standard mystery list of 2,315 words and guess list of 12,972 words. 

\begin{figure}[ht]
\begin{center}
\centerline{{\includegraphics[width=0.8\columnwidth]{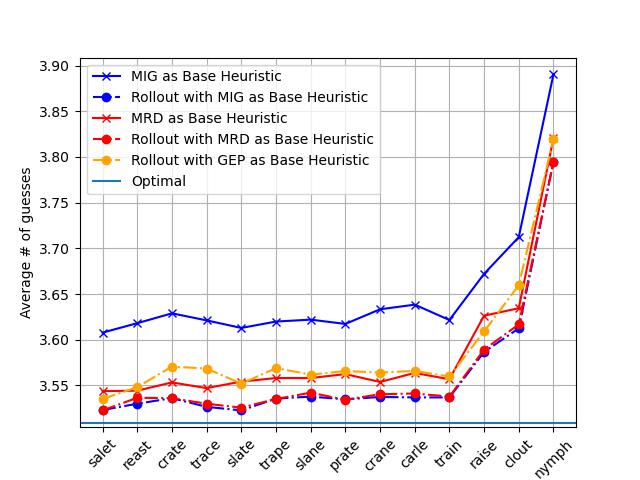}}}
\caption{Comparing base heuristic results with our rollout approach in the hard mode. The optimal score shown assumes the best opening guess word  (\texttt{salet}), and is equal to 3.5084 (cf.\ Table 1).}
\label{fig:hard_mode_chart}
\end{center}
\end{figure}

We provide a comparison of the performance of the three base heuristics, with and without rollout. In Table \ref{tab:max_info_gain} we give our results for the MIG heuristic for both the easy and the hard mode of the game along with optimal scores as shown by [Sel22] and [BeP22], and in Table \ref{tab:mrd_gep_combined}  we give our results for the MRD and GEP heuristics in just the hard mode.\footnote{Due to a significantly higher time taken for a single opening word to solve all the mystery words using the MRD and GEP heuristics, we only show the comparison for the hard mode for these two heuristics.} In both tables, we show the score for each opening word, averaged over all the 2,315 mystery words. We also compare the base heuristic and rollout performances in Fig.\ \ref{fig:hard_mode_chart}.

\begin{table}[ht]
\caption{Results using ``most rapid decrease" (MRD) and ``greatest expected probability" (GEP) as base heuristics for our rollout approach, in hard mode.}
\label{tab:mrd_gep_combined}
\resizebox{\textwidth}{!}{%
\begin{tabular}{|c|c|c|c|c|}
\hline
\textbf{Opening Word}   & \textbf{\begin{tabular}[c]{@{}c@{}}MRD as \\ Base Heuristic\end{tabular}} & \textbf{\begin{tabular}[c]{@{}c@{}}Rollout with\\ MRD as \\ Base Heuristic\end{tabular}} & \textbf{\begin{tabular}[c]{@{}c@{}}GEP as \\ Base Heuristic\end{tabular}} & \textbf{\begin{tabular}[c]{@{}c@{}}Rollout with\\ GEP as \\ Base Heuristic\end{tabular}} \\ \hline
\textbf{salet} & 3.5438                                                                    & 3.5227                                                                                   & 5.8674                                                                    & 3.5352                                                                                   \\ \hline
\textbf{reast} & 3.5443                                                                    & 3.5365                                                                                   & 5.9244                                                                    & 3.5481                                                                                   \\ \hline
\textbf{crate} & 3.5533                                                                    & 3.5361                                                                                   & 5.8998                                                                    & 3.5706                                                                                   \\ \hline
\textbf{trace} & 3.5471                                                                    & 3.53                                                                                     & 5.8695                                                                    & 3.5685                                                                                   \\ \hline
\textbf{slate} & 3.5542                                                                    & 3.5257                                                                                   & 5.8445                                                                    & 3.552                                                                                    \\ \hline
\textbf{trape} & 3.5581                                                                    & 3.5352                                                                                   & 5.8479                                                                    & 3.5689                                                                                   \\ \hline
\textbf{slane} & 3.5581                                                                    & 3.5421                                                                                   & 5.9158                                                                    & 3.5619                                                                                   \\ \hline
\textbf{prate} & 3.5624                                                                    & 3.5343                                                                                   & 5.8462                                                                    & 3.5658                                                                                   \\ \hline
\textbf{crane} & 3.5538                                                                    & 3.5404                                                                                   & 5.9935                                                                    & 3.5641                                                                                   \\ \hline
\textbf{carle} & 3.5637                                                                    & 3.5412                                                                                   & 5.9788                                                                    & 3.5659                                                                                   \\ \hline
\textbf{train} & 3.5568                                                                    & 3.5378                                                                                   & 5.8907                                                                    & 3.5598                                                                                   \\ \hline
\textbf{raise} & 3.6263                                                                    & 3.5892                                                                                   & 6.206                                                                     & 3.6091                                                                                   \\ \hline
\textbf{clout} & 3.6345                                                                    & 3.6168                                                                                   & 5.9974                                                                    & 3.6596                                                                                   \\ \hline
\end{tabular}%
}
\end{table}

It can be seen that our rollout approach consistently performs substantially better than the corresponding base heuristics. We also note that even for the GEP heuristic that initially does not perform well, our rollout approach still improves the performance significantly to near-optimal level. \old{It imperatively shows how one-step look-ahead minimization search leads to a substantial improvement in terms of cost (here, cost can be interpreted as the average score which we try to minimize) and computationally too (as we select only the top 10 next guess words using the base heuristic and select the best one using their approximate Q-factors).}

\section{A Reinforcement Learning Approach Towards POMDP and Adaptive Control}
\label{sec4:pomdp}

In this section, we focus on a problem of adaptive control of a deterministic discrete-time system that involves a state, a control, and an unknown parameter. This problem includes as special cases a variety of sequential search problems, including the Wordle puzzle. We will formulate the problem within a POMDP framework, a classical approach in adaptive control (see e.g., [Ber17], Section 6.8, and the references quoted there). We will then describe the corresponding DP algorithm, and reinforcement learning schemes that are based on approximation in value space. One of these schemes involves rollout ideas and contains as a special case the rollout algorithm for Wordle, which we have described in Section \ref{sec2:formulation}. 

Let us denote by $x_k$ and $u_k$ the  state and control of the system at time $k$, respectively. The control $u_k$ is selected from a known subset $U(x_k)$ of a control space, which may depend on the state $x_k$. Let us also denote by $\theta$ the unknown system parameter. We assume that $\theta$ stays fixed over time, at one of $m$ known values $\theta^1,\ldots,\theta^m$:
$$\theta\in\{\theta^1,\ldots,\theta^m\}.$$
The state $x_k$ is assumed to be perfectly observed by the controller at each time $k$, and evolves according to a system equation
\begin{equation} \label{eq-system}
x_{k+1}=f(x_k,\theta,u_k),
\end{equation}
over a horizon of $N$ time periods.

The a priori probability distribution of $\theta$ is given, and is updated based on the (perfectly) observed values of $x_k$ and the applied controls $u_k$. In particular, we assume that  the following information vector,
$$I_k=\{x_0,\ldots,x_k,u_0,\ldots,u_{k-1}\},$$
 is available at time $k$, and is used to compute the conditional probabilities
\begin{equation}\label{eq-beliefprobs}
b_{k,i}=P\{ \theta=\theta^i\mid I_k\},\qquad i=1,\ldots,m.
\end{equation}
These  probabilities form a vector 
$$b_k=(b_{k,1},\ldots,b_{k,m}),$$
which together with the perfectly observed state $x_k$, form the pair $(x_k,b_k)$ that is commonly called the {\it belief state} of the system at time $k$.

Following the choice of $u_k$, a cost $g(x_k,\theta,u_k)$ is incurred, and we wish to choose controls to minimize the sum of the incurred costs over a given number of stages $N$.
This is a classical  adaptive optimal control problem, involving simultaneous estimation of the system parameter $\theta$ and selection of the controls $u_k$. In the control theory literature, this problem has been addressed since the 1960s under the names {\it dual control} or {\it indirect adaptive optimal control\/}, and its optimal solution is known to be notoriously difficult. 

The introduction of the belief state is a hallmark of the classical partially observed Markovian decision problem (POMDP), which is among the most challenging DP problems, and in practice requires the use of approximations for suboptimal solution. The POMDP approach to adaptive control is part of the folklore of control theory. For example, it is discussed in detail in the survey by Kumar [Kum85], which gives many related references.
Our methodology of this paper and its extensions to stochastic systems, discussed in Section 5, may be viewed either as an RL approach for  adaptive control or as a suboptimal control approach for POMDP of a special type, whereby part of the state, namely $x_k$, is perfectly observed as it dynamically evolves, whereas another part of the state, namely $\theta$, is  fixed but unobservable, except indirectly through its influence on $x_k$. 

We note that an interesting application of our adaptive control framework arises in search problems, where $\theta$ specifies the locations of one or more hidden objects of interest within a given space. These locations gradually become known through the use of  sequential observations. We wish to determine the locations of the hidden objects with minimum total observation cost. The conceptual and methodological connection between adaptive control, sequential search problems, and DP is well-known, and has been explored in the literature since the 1950s. Generally, sequential search may be viewed as a special case of adaptive control, which in turn may be viewed as a special case of DP with partial state observations.

The artificial intelligence view of RL also has a  connection with adaptive control, and places strong emphasis on unknown problem environments. In particular, to quote from the book by Sutton and Barto  [SuB18],  ``learning from interaction with the environment is  a foundational idea underlying nearly all theories of learning and intelligence."  The idea of interaction with the environment is typically connected with the idea of identifying the environment characteristics. In control theory this is often viewed as part of the system identification methodology, which aims to construct parametric models of dynamic systems using data. The system identification process is often combined with the control process to deal with unknown or changing problem parameters, similar to the methodology described in this section. 

Note, that according to the classical methodology of POMDP (see e.g., [Ber17], Chapter 4), the belief component $b_{k+1}$ is determined by the belief state $(x_k,b_k)$, the control $u_k$, and the observation obtained at time $k+1$, i.e., $x_{k+1}$. Thus $b_k$ can be updated according to an equation of the form
\begin{equation} \label{eq-estimator}
b_{k+1}=B_k(x_k,b_k,u_k,x_{k+1}),
\end{equation}
where $B_k$ is an appropriate function, 
which can be viewed as a recursive estimator of $\theta$. There are several approaches to implement this estimator (perhaps with some approximation error), including the use of Bayes' rule and the simulation-based method of particle filtering, but we will not discuss this issue further in this paper. 

To place the Wordle puzzle within our adaptive control context, we view the mystery word $\theta$ as the  unknown system parameter, and we view the list of the mystery words as the set   $\{\theta^i\mid i=1,\ldots,m\}$ of possible values for $\theta$. The initial distribution of $\theta$ is uniform over the list of the mystery words, as is the case in the New York Times version of the puzzle. It can then be shown that the belief distribution $b_k$ at stage $k$ continues to be uniform over the list of eligible mystery words (those that have not been excluded by the preceding word guesses). This is an important simplification, which obviates the need for the estimator \eqref{eq-estimator}. An important consequence is that we may use as state $x_k$ the list of eligible mystery words at stage $k$, which evolves according to a deterministic system equation $x_{k+1}=f(x_k,u_k)$, with $u_k$ being the guess word at stage $k$. The function $f$ defining this system equation is specified by the rules for shrinking the mystery list, as described in Section 2. 

\subsection{The Exact DP Algorithm and its Approximation in Value Space}

We will now describe an exact DP algorithm for the adaptive control problem of this section. The algorithm operates in the space of information vectors $I_k$. In particular, we denote by $J_k(I_k)$ the optimal cost starting at information vector $I_k$ at time $k$. This vector evolves over a finite number of stages $N$ according to the equation
\begin{equation} \label{eq-infoequation}
I_{k+1}=(I_k,x_{k+1},u_k)=\big(I_k,f(x_k,\theta,u_k),u_k\big),\qquad k=0,\ldots,N-1.
\end{equation}
We may view this equation as a dynamic system whose state is $I_k$, the control is $u_k$, and $x_{k+1}$ is a stochastic ``disturbance" whose probability distribution depends on $(I_k,u_k)$. This is a standard formulation in POMDP. It admits a DP algorithm that takes the form
\begin{equation} \label{eq-dpalgorithmorig}
J_k^*(I_k)=\min_{u_k\in U(x_k)}E_{\theta}\Big\{g(x_k,\theta,u_k)+J_{k+1}^*\big(I_k,f(x_k,\theta,u_k),u_k\big)\mid I_k,u_k\Big\},
\end{equation}
for $k=0,\ldots,N-1$, with 
$$J_N^*(I_N)=g_N(x_N),$$
where we use $E_{\theta}\{\cdot\mid I_k,u_k\}$ to denote expected value over $\theta$, conditioned on $I_k$ and $u_k$; see e.g., the DP textbook [Ber17], Section 4.1. The algorithm produces the optimal costs $J_k^*(I_k)$ starting at information vector $I_k$ at time $k$. The optimal value of the problem is $J_0^*(I_0)$, where $I_0$ is the initial information vector, i.e., the state $x_0$. 

We can rewrite this DP algorithm in terms of the conditional belief probabilities $b_{k,i}$ as
\begin{equation} \label{eq-dpalgorithm}
J_k^*(I_k)=\min_{u_k\in U(x_k)}\sum_{i=1}^m b_{k,i}\Big\{g(x_k,\theta^i,u_k)+J_{k+1}^*\big(I_k,f(x_k,\theta^i,u_k),u_k\big)\Big\}.
\end{equation}
The control applied by the optimal policy is given by
\begin{equation} \label{eq-optpolicy}
u_k^*\in\arg\min_{u_k\in U(x_k)}\sum_{i=1}^m b_{k,i}\Big\{g(x_k,\theta^i,u_k)+J_{k+1}^*\big(I_k,f(x_k,\theta^i,u_k),u_k\big)\big)\Big\}.
\end{equation}
Note that the control $u_k^*$ generated by the preceding minimization is a function of $I_k$, since the belief probabilities $b_{k,i}$ are themselves determined by $I_k$, cf.\ Eq.\ \eqref{eq-beliefprobs}.

On the other hand, the algorithm \eqref{eq-dpalgorithm} is typically very hard to implement, because of the dependence of $J_{k+1}^*$ on the entire information vector $I_{k+1}$, which expands in size according to
Eq.\ \eqref{eq-infoequation}.
To address this implementation difficulty, we may use approximation in value space, based on replacing $J_{k+1}^*$ in the DP algorithm \eqref{eq-dpalgorithmorig}\  with some function $\tilde J_{k+1}$ that can be obtained (either off-line or on-line) with a tractable computation. The corresponding approximation in value space scheme with one-step lookahead minimization is given by
\begin{equation} \label{eq-lookschemegeneral}
\tilde u_k\in\arg\min_{u_k\in U(x_k)}\sum_{i=1}^m b_{k,i}\Big\{g(x_k,\theta^i,u_k)+\tilde J_{k+1}\big(I_k,f(x_k,\theta^i,u_k),u_k\big)\Big\}.
\end{equation}

A special type of approximation possibility is based on the use of the optimal cost function that corresponds to each parameter value $\theta^i$,
\begin{equation} \label{eq-optcosts}
J_{k+1}^i(x_{k+1}),\qquad i=1,\ldots,m.
\end{equation}
Here, $J_{k+1}^i(x_{k+1})$ is the optimal cost that would be obtained starting from state $x_{k+1}$ under an ``oracle" assumption, namely that $\theta$ is known to be equal to $\theta^i$; this corresponds to a perfect state information problem, which may be solvable under favorable circumstances.\footnote{In favorable special cases, the  costs $J_{k+1}^i(x_{k+1})$ may be easily calculated in closed form. Still, however, even in such cases the calculation of the belief probabilities $b_{k,i}$ may not be simple, and may require the use of a system identification algorithm.} The corresponding approximation in value space scheme with one-step lookahead minimization is given by
\begin{equation} \label{eq-lookschemeidependence}
\tilde u_k\in\arg\min_{u_k\in U(x_k)}\sum_{i=1}^m b_{k,i}\Big\{g(x_k,\theta^i,u_k)+J_{k+1}^i\big(f(x_k,\theta^i,u_k)\big)\Big\}.
\end{equation}

In the case where the horizon is sufficiently long, it is reasonable to expect that the estimate of the parameter $\theta$ improves over time, and that with a suitable estimation scheme, it converges  to the correct value of $\theta$, i.e., 
$$
 \lim_{k\to\infty} b_{k,i}=\begin{cases}
1& \text{if $\theta^i=\theta$,}\\
0& \text{if $\theta^i\ne\theta$.}
\end{cases}
$$
Then, asymptotically, the expression that is minimized in Eq.\ \eqref{eq-lookschemeidependence} is the same as the expression minimized in the Bellman equation that corresponds to the correct parameter $\theta$. Thus the generated one-step lookahead controls $\tilde u_k$ obtained from Eq.\ \eqref{eq-lookschemeidependence} are typically optimal in an asymptotic sense. Schemes of this type have been discussed in the adaptive control literature since the 70s; see e.g., Mandl [Man74], Doshi and Shreve [DoS80], Kumar and Lin [KuL82], Kumar [Kum85]. Moreover, some of the pitfalls of performing parameter identification while simultaneously applying adaptive control have been described by Borkar and Varaiya [BoV79], and by Kumar [Kum83]; see [Ber17], Section 6.8, for a related discussion.

Another, less computationally demanding approach is to use an approximation $\tilde J_{k+1}^i$ in place of the function $J_{k+1}^i$:
$$\tilde J_{k+1}^i(x_{k+1})\approx J_{k+1}^i(x_{k+1})\qquad \text{for every $x_{k+1}$}.$$
The corresponding approximation in value space scheme with one-step lookahead minimization is given by
\begin{equation} \label{eq-lookscheme}
\tilde u_k\in\arg\min_{u_k\in U(x_k)}\sum_{i=1}^m b_{k,i}\Big\{g(x_k,\theta^i,u_k)+\tilde J_{k+1}^i\big(f(x_k,\theta^i,u_k)\big)\Big\};
\end{equation}
cf.\ Eq.\ \eqref{eq-lookschemeidependence}. A simpler version of this approach is to use the same cost function approximation $\tilde J_{k+1}$ for every $i$; i.e.,
$$\tilde J_{k+1}^i(x_{k+1})=\tilde J_{k+1}(x_{k+1}),\qquad \text{for every $i$ and $x_{k+1}$}.$$
 However, the dependence of $\tilde J_{k+1}$ on $i$ may be useful in some contexts where differences in the value of $i$ may have a radical effect on the qualitative character of the system equation.

\subsection{Rollout}

We will now discuss a rollout algorithm that is similar in structure to other rollout algorithms that have been considered in the literature; see the book [Ber20], which provides extensive discussion and connections to other methodologies in RL, model predictive control, and discrete optimization. Generally, rollout algorithms use as cost function approximations the cost functions of given policies, called base policies (or base heuristics). Rollout algorithms are related to the method of policy iteration, and in fact they may be viewed as a single policy iteration (perhaps with some approximations). They generally aspire to a policy improvement property, inherited from their connection to policy iteration, namely that they perform at least as well as the base policy.

The rollout algorithm that we propose in this paper  can be viewed as a special case of the approximation in value scheme of Eq.\ \eqref{eq-lookscheme}, with the cost functions $J_{k+1,\pi^i}^i(x_{k+1})$ of given  policies $\pi^i$ used as the cost function approximations $\tilde J_{k+1}^i(x_{k+1})$.  In this case Eq.\ \eqref{eq-lookscheme} takes the form 
\begin{equation} \label{eq-policylookscheme}
\tilde u_k\in\arg\min_{u_k\in U(x_k)}\sum_{i=1}^m b_{k,i}\Big\{g(x_k,\theta^i,u_k)+J_{k+1,\pi^i}^i\big(f(x_k,\theta^i,u_k)\big)\Big\},
\end{equation}
and has the character of a rollout algorithm, with $\pi^i$, $i=1,\ldots,m$, being known base policies, which at stage $k$ apply control that depends only on $x_k$. 

Here, the term
$$J_{k+1,\pi^i}^i\big(f(x_k,\theta^i,u_k)\big)$$
in Eq.\ \eqref{eq-policylookscheme}\ is the cost of the base policy $\pi^i$, starting from the next state 
$$x_{k+1}=f(x_k,\theta^i,u_k),$$
under the ``oracle" assumption that $\theta$ will stay fixed at the value $\theta=\theta^i$.
Thus $J_{k+1,\pi^i}^i\big(f(x_k,\theta^i,u_k)\big)$ is readily computed by  
deterministic propagation of the state of the system \eqref{eq-system}, starting from $x_{k+1}=f(x_k,\theta^i,u_k)$ up to the end of the horizon, using the base policy $\pi^i$ and assuming that $\theta$ is fixed at the value $\theta^i$. Moreover, the needed values of $J_{\pi^i}^i$ in Eq.\ \eqref{eq-policylookscheme} are uncoupled over $i$, and their computation can be done in parallel for different $i$. As in the case of approximation in value space, a simpler possibility is to use  the same policy for all $i$, i.e., $\pi^i=\pi$ for all $i$, where  $\pi$ is some policy.

An example of the above suboptimal control selection approach is the MIG heuristic in the Wordle context, where:
\begin{itemize}
\item [(a)]  $x_k$ is the list  of eligible mystery words at stage $k$.
\item [(b)] $b_k$ is the uniform distribution over $x_k$, so $b_{k,i}$ equal to either 0 or to $1/m_k$, where $m_k$ is the number of words in $x_k$.
\item [(c)]  $u_k$ is the $k$th guess word selected, while $U(x_k)$ is the list $U$ of guess words.
\item [(d)] $g(x_k,\theta^i,u_k)=1$.
\item [(e)] $\tilde J_{k+1}$ in Eq.\ \eqref{eq-lookschemegeneral} is a function of $x_k$ and $u_k$ only, and the value $\tilde J_{k+1}(x_k,u_k)$ is equal to the entropy $E(x_k,u_k)$ of the distribution that corresponds to $x_k$ and $u_k$.
\end{itemize}
The MIG heuristic selects at $x_k$ the guess word that  minimizes $E(x_k,u_k)$ over $u_k$.

The corresponding rollout algorithm takes the form
\begin{equation} \label{eq-rollout}
\tilde u_k\in \arg\min_{u_k\in U}\sum_{\{i\mid \theta^i\in x_k\}}\tilde J_{k+1}^{i}\big(f(x_k,\theta^i,u_k)\big),
\end{equation}
where as earlier:
\begin{itemize}
\item [(a)]  $x_k$ is the list  of eligible mystery words at stage $k$.
\item [(b)] $u_k$ is the $k$th guess word to be selected from the list $U$ of guess words.
\item [(c)] The value $\tilde J_{k+1}^{i}\big(f(x_k,\theta^i,u_k)\big)$ is the cost that would be incurred by the MIG heuristic under the ``oracle" assumption that the mystery word is $\theta^i$, and starting from the list of eligible mystery words $f(x_k,\theta^i,u_k)$ at stage $k+1$.
\end{itemize}
Thus when the current list of eligible words is $x_k$, the rollout algorithm computes for each eligible mystery word $\theta^i$ and for each guess word $u_k$, the corresponding costs $\tilde J_{k+1}^{i}\big(f(x_k,\theta^i,u_k)\big)$ of the information-based heuristic starting from $f(x_k,\theta^i,u_k)$. It then uses the guess word $\tilde u_k$ that minimizes the sum of these costs, as per Eq.\ \eqref{eq-rollout}.

\section{Generalizations and Extensions}
\label{sec5:extensions}

In this section we discuss variations and extensions of our algorithmic approach. The book [Ber22], Section 6.7, describes a generalization of our adaptive control approach to a stochastic system, and will be summarized next.

\subsection{A POMDP/Adaptive Control Approach for Stochastic Systems}

In a stochastic version of the formulation of Section 3, the deterministic system equation \eqref{eq-system} is replaced by a stochastic equation of the form 
$$x_{k+1}=f(x_k,\theta,u_k,w_k),$$
where $u_k$ is the control at time $k$, selected from a set $U(x_k)$, and $w_k$ is a random disturbance with given probability distribution that depends on $(x_k,\theta,u_k)$. All other aspects of the problem are unchanged, except for the cost per stage, $g(x_k,\theta,u_k,w_k)$, which now depends on the random disturbance $w_k$.

The exact DP algorithm that operates in the space of information vectors $I_k$ is now given by
\begin{dmath*}
J_k^*(I_k)=\min_{u_k\in U(x_k)}E_{\theta,w_k}\Big\{g(x_k,\theta,u_k,w_k)+{J_{k+1}^*\big(I_k,f(x_k,\theta,u_k,w_k),u_k\big)\mid I_k,u_k\Big\}}
\end{dmath*}
for $k=0,\ldots,N-1$, with $J_N^*(I_N)=g_N(x_N)$; see e.g., the DP textbook [Ber17], Section 4.1. 

By using the law of iterated expectations, 
$$E_{\theta,w_k}\{\cdot\mid I_k,u_k\}=E_\theta\big\{E_{w_k}\{\cdot\mid I_k,\theta,u_k\}\mid I_k,u_k\big\},$$
we can rewrite this DP algorithm as
\begin{align*}
J_k^*(I_k)=\min_{u_k\in U(x_k)}\sum_{i=1}^m b_{k,i}E_{w_k}\Big\{&g(x_k,\theta^i,u_k,w_k)+\cr
&{J_{k+1}^*\big(I_k,f(x_k,\theta^i,u_k,w_k),u_k\big)\mid I_k,\theta^i,u_k\Big\}}.
\end{align*}
The summation over $i$ above represents the expected value of $\theta$ conditioned on $I_k$ and $u_k$. 

Similar to Section \ref{sec4:pomdp}, we may use an approximation to the optimal cost function corresponding to each parameter value $\theta^i$,
$$\tilde J_{k+1}^i(x_{k+1}),\qquad i=1,\ldots,m.$$
The corresponding approximation in value space scheme is given by
\begin{align*}
\tilde u_k\in\arg\min_{u_k\in U(x_k)}\sum_{i=1}^m b_{k,i}E_{w_k}\Big\{&g(x_k,\theta^i,u_k,w_k)+\cr
&{\tilde J_{k+1}^i\big(f(x_k,\theta^i,u_k,w_k)\big)\mid x_k,\theta^i,u_k\Big\}}.
\end{align*}
A simpler version of this approach is to use the same function $\tilde J_{k+1}^i$ for every $i$. 

The rollout algorithm of Section \ref{sec4:pomdp} may be similarly extended to the stochastic case of this section. We refer to the book [Ber22], Secton 6.7 for further discussion.

\subsection{More Challenging Versions of the Wordle Puzzle}

The rollout approach also applies to several variations of the Wordle puzzle. Such variations may include for example a larger length of mystery words, and/or a non-uniform distribution over the initial list of mystery words. For comparison purposes, we have implemented a 6-letter version of Wordle, where the mystery and guess lists consist of 6-letter words. We use the 6-letter word list obtained from the Natural Language Corpus\footnote{\url{http://norvig.com/ngrams/}}, and select 12,972 words for our guess word list, assuming a random distribution over the entire list of words. From this guess word list, we again randomly sample 2,315 mystery words to create our mystery word list for the 6-letter Wordle experiments.

\begin{table}[ht]
\caption{Results for 6-letter Wordle using maximum information gain (MIG) as base heuristic and the corresponding  rollout algorithm.}
\label{tab:6letter}
\resizebox{\textwidth}{!}{%
\begin{tabular}{|c|cc|cc|}
\hline
\multirow{2}{*}{\textbf{Opening Word}} & \multicolumn{2}{c|}{\textbf{Easy Mode}}                                                                                                                                                   & \multicolumn{2}{c|}{\textbf{Hard Mode}}                                                                                                                                                   \\ \cline{2-5} 
                                       & \multicolumn{1}{c|}{\textbf{\begin{tabular}[c]{@{}c@{}}MIG as \\ Base Heuristic\end{tabular}}} & \textbf{\begin{tabular}[c]{@{}c@{}}Rollout with\\ MIG as \\ Base Heuristic\end{tabular}} & \multicolumn{1}{c|}{\textbf{\begin{tabular}[c]{@{}c@{}}MIG as \\ Base Heuristic\end{tabular}}} & \textbf{\begin{tabular}[c]{@{}c@{}}Rollout with\\ MIG as \\ Base Heuristic\end{tabular}} \\ \hline
\textbf{ambros}                        & \multicolumn{1}{c|}{3.3235}                                                                    & 3.1637                                                                                   & \multicolumn{1}{c|}{3.3024}                                                                    & 3.1918                                                                                   \\ \hline
\textbf{rabies}                        & \multicolumn{1}{c|}{3.2346}                                                                    & 3.0898                                                                                   & \multicolumn{1}{c|}{3.2294}                                                                    & 3.1279                                                                                   \\ \hline
\textbf{tances}                        & \multicolumn{1}{c|}{3.2056}                                                                    & 3.0769                                                                                   & \multicolumn{1}{c|}{3.1978}                                                                    & 3.1041                                                                                   \\ \hline
\end{tabular}%
}
\end{table}

In Table \ref{tab:6letter}, we provide some sample experimental results using the MIG base heuristic. Naturally, when passing from 5-letter Wordle to 6-letter Wordle, it may become practically infeasible to compute the optimal solution given the exponential increase in the search space. However, using the rollout approach, we are able to solve all the mystery word games with an average computation time in the order of seconds, while showing a significant improvement over the MIG base heuristic once again.


\section{Concluding Remarks}
\label{sec6:conclusion}

In this paper we have described a rollout algorithm for solving the Wordle puzzle. Our computational results show that the performance of our algorithm is very close to optimal, and much better than the ones of the three different base heuristics that we have used. Moreover, our rollout approach is capable of addressing extensions of Wordle, as well as types of search and adaptive control problems for which exact solution by DP is impractical. 

The most challenging variant of Wordle appears to be one where the probability distribution of selection of the mystery word is non-uniform; e.g., when mystery words are selected consistently with their frequency of usage within some domain. It appears impossible to solve the puzzle optimally in this case, in the absence of special assumptions. For the same variant of Wordle, the main additional computation required by the rollout approach is the updating of the belief distribution over the mystery list, rather than updating the current mystery list [this is required to compute the average Q-factor of a guess word, in an analog of Eq.\ (\ref{eq:qfactor})]. This can be done analytically or more likely by using simulation-based methods such as particle filtering. Testing this approach computationally is an interesting subject for further research, either in the context of Wordle, or more generally in the context of the adaptive control problem of Section  \ref{sec4:pomdp}.

An important direction for further research is the use of our methodology in automated planning. For example, special types of POMDP involving a fully observable state component, and a constant partially observable component, have been investigated in a number of  works on planning. In particular, the papers [RaS14, Ran15a, Ran15b] and [Cha20] discuss multiple-environment Markov Decision Processes. These processes are described as Markov Decision Processes equipped with multiple probabilistic transition functions, which represent the various possible unknown environments. [Cha20] describes how this special structure can be exploited to facilitate the computational solution, and notes applications in recommender systems and multi-environment robot navigation.  Moreover, in addition to on-line POMDP planning works, there has been extensive research that deals with on-line and off-line planning methods for POMDP; see e.g., the book [GeB13], the papers [Bry06, Mal14, Mui14], and the tutorial survey [BrK07]. 

In this connection, it is also worth noting that rollout may be viewed as an on-line search method, which uses the cost function of the base policy as a heuristic function, in the terminology of on-line heuristic search (e.g., the books [Pea84, EdS11, GeB13]). However, in the rollout algorithm the heuristic function is not admissible, i.e., it is not an underestimate of the optimal cost function,  as in A$^*$, POMDP search, and related methods. Instead, being the cost function of a policy, it is an {\it overestimate} of the optimal cost function. As a result, it does not offer a guarantee of asymptotically optimal  performance, only a guarantee of cost improvement over the base policy. However, analytical insights (based on the Newton step interpretation noted earlier), as well as extensive computational practice (consistent with the computations reported in this paper) suggest that this cost improvement is typically substantial and often dramatic; see the discussion and the references to case studies given in the books [Ber19, Ber20].

\section*{Bibliography}

\def\ref{\hskip0pc\vskip2pt}

\ref [BeP22] Bertsimas D. and Paskov A., 2022.\ ``An Exact and Interpretable Solution to Wordle." Available at
\href{https://scholar.google.com/scholar?hl=en&as_sdt=0\%2C3&q=An+Exact+and+Interpretable+Solution+to+Wordle&btnG=&oq=an+}{URL}. Preprint, received June 2022. (Accessed: 14 November 2022). 

\ref[Ber17] Bertsekas, D.\ P., 2017.\
Dynamic Programming and Optimal Control, Vol.\ I,  4th Ed., Athena Scientific, Belmont, MA.

\ref[Ber19] Bertsekas, D.\ P., 2019.\ Reinforcement Learning and Optimal Control, Athena Scientific, Belmont, MA.

\ref[Ber20] Bertsekas, D.\ P., 2020.\
Rollout, Policy Iteration, and Distributed Reinforcement Learning, Athena Scientific, Belmont, MA.

\ref[Ber22] Bertsekas, D.\ P., 2022.\ Lessons from AlphaZero for Optimal, Model Predictive, and Adaptive Control, Athena Scientific, Belmont, MA.

\ref[BeT96] Bertsekas, D.\ P. and Tsitsiklis, J.\ N., 1996. Neuro-dynamic programming. Athena Scientific.

\ref [Bon22] Bonthron, M., 2022.\ ``Rank One Approximation as a Strategy for Wordle," arXiv:2204.06324.

\ref [BoV79] Borkar, V., and Varaiya, P.\ P., 1979.\  ``Adaptive Control of
Markov Chains, I:  Finite Parameter Set," IEEE Trans.\ Automatic Control, Vol.\
AC-24, pp.\ 953-958.

\ref [Bry06] Bryce, D., Kambhampati, S. and Smith, D.E., 2006. ``Planning graph heuristics for belief space search." Journal of Artificial Intelligence Research, 26, pp.35-99.

\ref [BrK07] Bryce, D. and Kambhampati, S., 2007. ``A tutorial on planning graph based reachability heuristics". AI Magazine, 28(1), pp.47-47.

\ref [Cha20] Chatterjee, K., Chmelík, M., Karkhanis, D., Novotný, P. and Royer, A., 2020, June. ``Multiple-Environment Markov Decision Processes: Efficient Analysis and Applications". In Proceedings of the International Conference on Automated Planning and Scheduling (Vol. 30, pp. 48-56).

\ref [DoS80] Doshi, B., and Shreve, S., 1980.\  ``Strong Consistency of a Modified
Maximum Likelihood Estimator for Controlled Markov Chains," J.\ of Applied
Probability, Vol.\ 17, pp.\ 726-734.

\ref [EdS11] Edelkamp, S. and Schrodl, S., 2011. Heuristic search: theory and applications. Elsevier.

\ref [GeB13] Geffner, H. and Bonet, B., 2013. A concise introduction to models and methods for automated planning. Synthesis Lectures on Artificial Intelligence and Machine Learning, 8(1), pp.1-141.

\ref [HoA22] Ho A., 2022.\ ``Solving Wordle with Reinforcement Learning." Available at 
\href{https://wandb.ai/andrewkho/wordle-solver/reports/Solving-Wordle-with-Reinforcement-Learning--VmlldzoxNTUzOTc4}{URL}. (Accessed: 14 November 2022).

\ref [KuL82] Kumar, P.\ R., and Lin, W., 1982.\  ``Optimal Adaptive Controllers for
Unknown Markov Chains," IEEE Trans.\ Automatic Control, Vol.\ AC-27, pp.\ 765-774.

\ref [Kum83] Kumar, P.\ R., 1983.\  ``Optimal Adaptive
Control of Linear\\-Quadratic-Gaussian Systems," SIAM J.\ on Control and Optimization,
Vol.\ 21, pp.\ 163-178.

\ref [Kum85] Kumar, P.\ R., 1985.\  ``A Survey of Some Results in Stochastic Adaptive
Control," SIAM J.\ on Control and Optimization, Vol.\ 23, pp.\ 329-380.

\ref [LoS22] Lokshtanov D., and Subercaseaux B., 2022.\ ``Wordle is NP-Hard," arXiv:2203.16713.

\ref [Mal14] Maliah, S., Brafman, R., Karpas, E. and Shani, G., 2014, May. ``Partially observable online contingent planning using landmark heuristics." In Twenty-Fourth International Conference on Automated Planning and Scheduling.

\ref[Man74] Mandl, P., 1974.\ ``Estimation and Control in Markov Chains," Advances in Applied Probability, Vol.\ 6, pp.\ 40-60.

\ref [Mui14] Muise, C., Belle, V. and McIlraith, S., 2014, June. ``Computing contingent plans via fully observable non-deterministic planning." In Proceedings of the AAAI Conference on Artificial Intelligence (Vol. 28, No. 1).

\ref [Pea84] Pearl, J., 1984. Heuristics: intelligent search strategies for computer problem solving. Addison-Wesley Longman Publishing Co., Inc..

\ref [Ran15a] Randour, M., Raskin, J.F. and Sankur, O., 2015, July. ``Percentile queries in multi-dimensional Markov decision processes". In International Conference on Computer Aided Verification (pp. 123-139). Springer, Cham.

\ref [Ran15b] Randour, M., Raskin, J.F. and Sankur, O., 2015, January.      ``Variations on the stochastic shortest path problem". In International Workshop on Verification, Model Checking, and Abstract Interpretation (pp. 1-18). Springer, Berlin, Heidelberg.

\ref [RaS14] Raskin, J.F. and Sankur, O., 2014. ``Multiple-environment Markov decision processes". arXiv preprint arXiv:1405.4733.

\ref [Ros22] Rosenbaum W., 2022.\ ``Finding a Winning Strategy for Wordle is NP-complete," arXiv:2204.04104.

\ref [Sel22] Selby A., 2022.\ ``The best strategies for Wordle (last edited on 17 March 2022)." Available at \href{https://sonorouschocolate.com/notes/index.php?title=The_best_strategies_for_Wordle}{URL}. (Accessed: 14 November 2022).

\ref [Sil22] de Silva, N., 2022.\ ``Selecting Seed Words for Wordle Using Character Statistics," arXiv:2202.03457.

\ref [SuB18] Sutton, R., and Barto, A.\  G.,  2018.\ Reinforcement Learning, 2nd Ed., MIT
Press, Cambridge, MA.

\ref [Sho22] Short, M.B., 2022.``Winning Wordle Wisely," arXiv preprint \\arXiv:2202.02148.

\ref [War22] Wardle J., 2022.\ ``Wordle is a love story." Available at \href{https://www.nytimes.com/2022/01/03/technology/wordle-word-game-creator.html}{URL}. (Accessed: 14 November 2022).

\end{document}